%% file: main.tex
\documentclass[10pt,twocolumn,letterpaper]{article}

\usepackage{iccv}
\usepackage{times}
\usepackage{epsfig}
\usepackage{graphicx}
\usepackage{amsmath}
\usepackage{amssymb}
\usepackage{booktabs}
\usepackage{arydshln}
\usepackage{multirow}
\usepackage{subfigure}
\usepackage[noend]{algpseudocode}
\usepackage{algorithm}
\usepackage[normalem]{ulem}
\usepackage{tablefootnote}
\usepackage[11pt]{moresize}

\usepackage[pagebackref=true,breaklinks=true,letterpaper=true,colorlinks,bookmarks=false]{hyperref}

\iccvfinalcopy 

\def\iccvPaperID{906} 

\ificcvfinal\fi
\begin{document}

\title{Visual Relationship Detection with \\Internal and External Linguistic Knowledge Distillation}

\author{Ruichi Yu, Ang Li, Vlad I. Morariu, Larry S. Davis\\
University of Maryland, College Park\\
{\tt\small \{richyu,angli,morariu,lsd\}@umiacs.umd.edu}
}

\maketitle
\newcommand\twotup[2]{$\langle #1,#2 \rangle$}
\newcommand\threetup[3]{$\langle #1,#2,#3\rangle$}
\newcommand\angcomment[1]{{\color{red} Comment[Ang]: #1}}
\newcommand\angdel[1]{{\color{blue}\sout{#1}}}


\begin{abstract}
Understanding the visual relationship between two objects involves identifying the \textit{subject}, the \textit{object}, and a predicate relating them.
We leverage the strong correlations between the predicate and the \twotup{subj}{obj} pair (both semantically and spatially) to predict predicates conditioned on the subjects and the objects. 
Modeling the three entities jointly more accurately reflects their relationships compared to modeling them independently, but it complicates learning since the semantic space of visual relationships is huge and training data is limited, especially for long-tail relationships that have few instances. 
To overcome this, we use knowledge of linguistic statistics to regularize visual model learning. 
We obtain linguistic knowledge by mining from both training annotations (internal knowledge) and publicly available text, \eg, Wikipedia (external knowledge), computing the conditional probability distribution of a predicate given a \twotup{subj}{obj} pair. As we train the visual model, we distill this knowledge into the deep model to achieve better generalization.
Our experimental results on the Visual Relationship Detection (VRD) and Visual Genome datasets suggest that with this linguistic knowledge distillation, our model outperforms the state-of-the-art methods significantly, especially when predicting unseen relationships (\eg, recall improved from 8.45\% to 19.17\% on VRD zero-shot testing set). 

\end{abstract}
\input{intro}

\input{relatedWork}
\input{K}

\input{Experiment}

\section{Conclusion}
We proposed a framework that distills linguistic knowledge into a deep neural network for visual relationship detection. We incorporated rich representations of a visual relationship in our deep model, and utilized a teacher-student distillation framework to help the data-driven model absorb internal (training annotations) and external (public text on the Internet) linguistic knowledge. Experiments on the VRD and the Visual Genome datasets show significant improvements in accuracy and generalization capability.

\section*{Acknowledgement}
The research was supported by the Office of Naval Research under Grant N000141612713: Visual Common Sense Reasoning for Multi-agent Activity Prediction and Recognition.

\newpage
{\small
\bibliographystyle{ieee}
\bibliography{egbib}
}

\end{document}

%% file: intro.tex
\section{Introduction}
Detecting visual relationships from images is a central problem in image understanding.
Relationships are commonly defined as tuples consisting of a subject (\textit{subj}), predicate (\textit{pred}) and object (\textit{obj}) \cite{nlp1,nlp2,nlp3}. \textit{Visual} relationships represent the visually observable interactions between subject and object \twotup{subj}{obj} pairs, such as \threetup{person}{ride}{horse} \cite{VRD}. 

\begin{figure}[!t]
\centering
  \includegraphics[width=\linewidth]{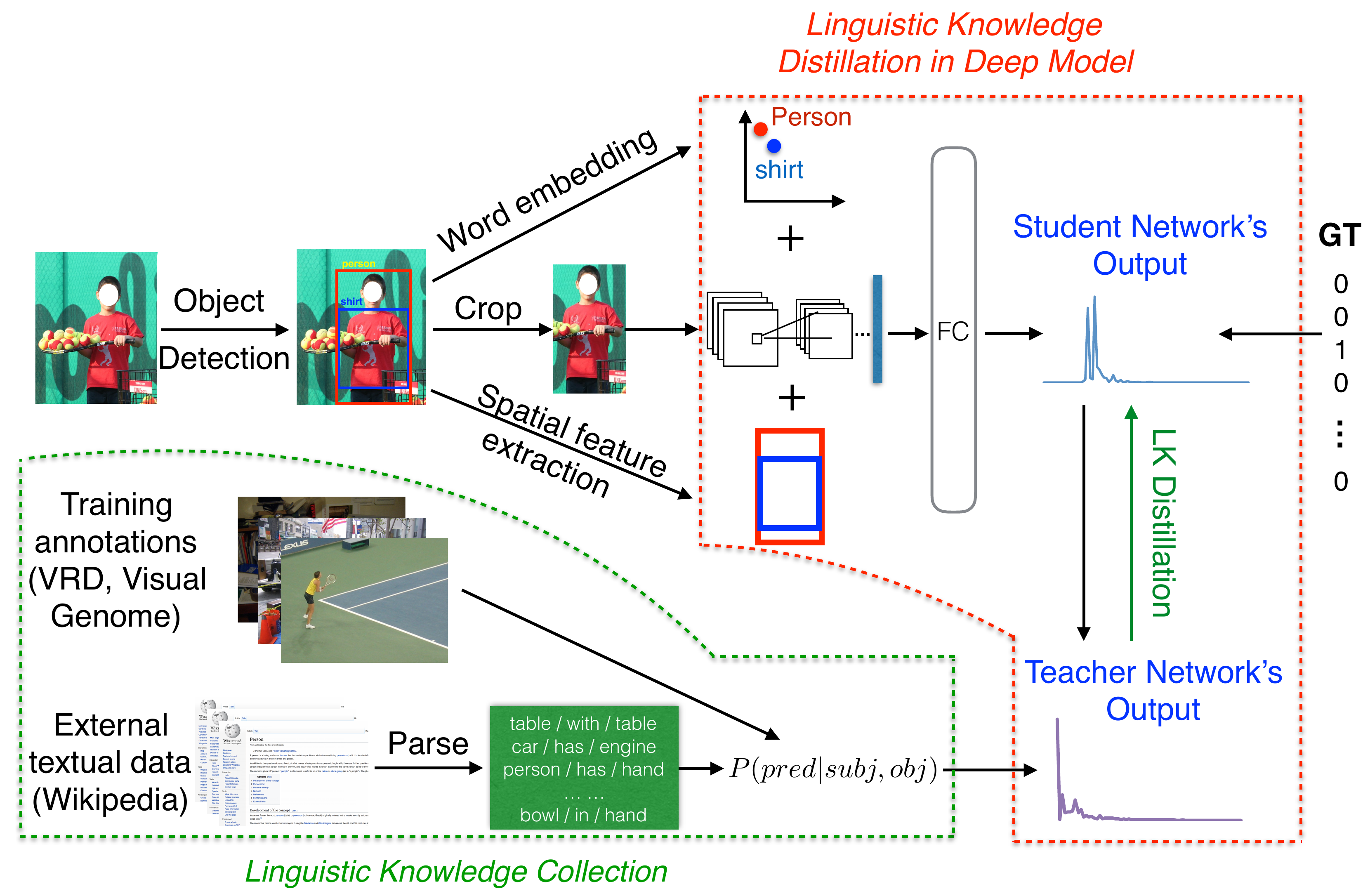}
  \caption{Linguistic Knowledge Distillation Framework. We extract linguistic knowledge from training annotations and a public text corpus (green box), then construct a teacher network to distill the knowledge into an end-to-end deep neural network (student) that predicts visual relationships from visual and semantic representations (red box). GT is the ground truth label and ``+" is the concatenation operator.}
\label{fig:system}
\end{figure}

Recently, Lu \etal \cite{VRD} introduce the visual relationship dataset (VRD) to study learning of a large number of visual relationships from images. 
Lu \etal predict the predicates independently from the subjects and objects, and use the product of their scores to predict relationships present in a given image using a linear model.
The results in \cite{VRD} suggest that predicates cannot be predicted reliably with a linear model that uses only visual cues, even when the ground truth categories and bounding boxes of the subject and object are given (\cite{VRD} reports Recall@100 of only 7.11\% for their visual prediction). Although the visual input analyzed by the CNN in \cite{VRD} includes the subject and object, predicates are predicted without any knowledge about the object categories present in the image or their relative locations. 
In contrast, we propose a probabilistic model to predict the predicate name jointly with the subject and object names and their relative spatial arrangement:
\begin{align}
\label{decompose}
P(R|I) =P(pred|I_{\text{union}},subj,obj)P(subj)P(obj).
\end{align}

While our method models visual relationships more accurately than \cite{VRD}, our model's parameter space is also enlarged because of the large variety of relationship tuples. This leads to the challenge of insufficient labeled image data. The straightforward---but very costly---solution is to collect and annotate a larger image dataset that can be used to train this model. Due to the long tail distribution of relationships, it is hard to collect enough training images for all relationships. To make the best use of available training images, we leverage linguistic knowledge (LK) to regularize the deep neural network. One way to obtain linguistic knowledge is to compute the conditional probabilities $P(pred|subj,obj)$ from the training annotations.

However, the number of \threetup{subj}{pred}{obj} combinations is too large for each triplet to be observed in a dataset of annotated images, so the internal statistics (\eg, statistics of the VRD dataset) only capture a small portion of the knowledge needed.
To address this long tail problem, we collect external linguistic knowledge ($P(pred|subj,obj)$) from public text on the Internet (Wikipedia).
This external knowledge consists of statistics about the words that humans commonly use to describe the relationship between subject and object pairs,  and importantly, it includes 
pairs unseen in our training data. Although the external knowledge is more general, it can be very noisy (\eg, due to errors in linguistic parsing). 

We make use of the internal and external knowledge in a teacher-student knowledge distillation framework \cite{eric1,eric2}, shown in Figure 1, where the output of the standard vision pipeline, called the \textit{student} network, is augmented with the output of a model that uses the linguistic knowledge to score solutions; their combination is called the \textit{teacher} network. 
The objective is formulated so that the student not only learns to predict the correct one-hot ground truth labels but also to mimic the teacher's soft belief between predicates.

Our main contribution is that we exploit the role of both visual and linguistic representations in visual relationship detection and use internal and external linguistic knowledge to regularize the learning process of an end-to-end deep neural network to significantly enhance its predictive power and generalization. We evaluate our method on the VRD \cite{VRD} and Visual Genome (VG) \cite{VG} datasets. 
Our experiments using Visual Genome show that while the improvements due to training set size are minimal, improvements due to the use of LK are large, implying that with current dataset sizes, it is more fruitful to incorporate other types knowledge (\eg, LK) than to increase the  visual dataset size---this is particularly promising because visual data is expensive to annotate and there exist many readily available large scale sources of knowledge that have not yet been fully leveraged for visual tasks.

%% file: relatedWork.tex
\section{Related Work}
\textbf{Knowledge Distillation in Deep Neural Networks:} Recent work has exploited the use of additional information (or ``knowledge") to help train deep neural networks (DNN) \cite{k1,k2,k3,hinton}. Hinton \etal \cite{hinton} proposed a framework to distill knowledge, in this case the predicted distribution, from a large network into a smaller network. Recently, Hu \etal proposed a teacher-student framework to distill massive knowledge  sources, including logic rules, into DNNs \cite{eric1,eric2}.

\textbf{Visual Relationship Detection:} Visual relationships represent the interactions between object pairs in images. Lu \etal \cite{VRD} formalized visual relationship prediction as a task and provided a dataset with a moderate number of relationships. Before \cite{VRD}, a large corpus of work had leveraged the interactions between objects (\eg object co-occurrence, spatial relationships) to improve visual tasks \cite{rich,o1,o2,o3,o4,o5,Ang}. 
To enable visual relationship detection on a large scale, Lu \emph{et al.} \cite{VRD} decomposed the prediction of a relationship into two individual parts: detecting objects and predicting predicates. Lu \emph{et al.} used the sub-image containing the union of two bounding boxes of object pairs as visual input to predict the predicates and utilized language priors, such as the similarity between relationships and the likelihood of a relationship in the training data, to augment the visual module. 

Plummer \emph{et al.} \cite{UIUC} grounded phrases in images by fusing  several visual features like appearance, size, bounding boxes, and linguistic cues (like adjectives that describe attribute information). Despite focusing on phrase localization rather than visual phrase detection, when evaluated on the VRD dataset,  \cite{UIUC} achieved comparable results with \cite{VRD}. Recently, there are several new attempts for visual relationship detection task:  Liang \etal \cite{RL} proposed to detect relationships and attributes within a reinforcement learning framework; Li \etal \cite{VIPCNN} trained an end-to-end system boost relationship detection through better object detection; Bo \etal \cite{Bo} detected relationships via a relational modeling framework.
We combine rich visual and linguistic representations in an end-to-end deep neural network that absorbs external linguistic knowledge using the teacher-student framework during the training process to enhance prediction and generalization.  Unlike \cite{VRD}, which detected objects independently from relationship prediction, we model objects and relationships jointly. Unlike \cite{VIPCNN,RL,Bo}, which do not use linguistic knowledge explicitly, we focus on predicting predicates using the linguistic knowledge that models correlations between predicates and \twotup{subj}{obj} pairs, especially for the long-tail relationships. Unlike \cite{hinton,eric1,eric2}, which used either the teacher or the student as their final output, we combine both teacher and student networks, as they each have their own advantages: the teacher outperforms in cases with sufficient training data, while the student generalizes to cases with few or no training examples (the zero-shot case).

%% file: K.tex
\section{Our Approach}
A straightforward way to predict relationship predicates is to train a CNN on the union of the two bounding boxes that contain the two objects of interest as the visual input, fuse semantic features (that encode the object categories) and spatial features (that encode the relative positions of the objects) with the CNN features  (that encode the appearance of the objects), and feed them into a fully connected (FC) layer to yield an end-to-end prediction framework. However, the number of \threetup{subj}{pred}{obj} tuples is very large and the parameter space of the end-to-end CNN would be huge. 
While the subject, predicate, and object are not statistically independent, a CNN would require a massive amount of data to discover the dependence structure while also learning the mapping from visual features to semantic relationships.
To avoid over-fitting and achieve better predictive power without increasing the amount of visual training data, additional information is needed to help regularize the training of the CNN.

Figure \ref{fig:system} summarizes our proposed model. Given an image, we extract three input components: the cropped images of the union of the two detected objects (BB-Union); the semantic object representations obtained from the object category confidence score distributions obtained from the detector; and the spatial features (SF) obtained from pairs of detected bounding boxes.
We concatenate VGG features, semantic object vectors, and the spatial feature vectors, then train another FC layer using the ground truth label (GT) and the linguistic knowledge to predict the predicate. Unlike \cite{VRD}, which used the VGG features to train a linear model, our training is end-to-end without fixing the VGG-net. Following \cite{eric1,eric2}, we call the data-driven model the ``student", and the linguistic knowledge regularized model the ``teacher".

\subsection{Linguistic Knowledge Distillation}
\subsubsection{Preliminary: Incorporating Knowledge in DNNs}
The idea of incorporating additional information in DNNs has been exploited recently \cite{hinton,eric1,eric2}. We adapted Hu \emph{et al.}'s teacher-student framework \cite{eric1,eric2} to distill linguistic knowledge in a data-driven model. The teacher network is constructed by optimizing the following criterion:
\begin{equation}
\label{optimization}
\min_{t \in T}\text{KL}(t(Y)|| s_{\phi}(Y|X)) - C \mathbb{E}_t[L(X,Y)],
\end{equation}
where $t(Y)$ and $s_{\phi}(Y|X)$ are the prediction results of the teacher and student networks; C is a balancing term; $\phi$ is the parameter set of the student network; $L(X,Y)$ is a general constraint function that has high values to reward the predictions that meet the constraints and penalize the others. $\text{KL}$ measures the KL-divergence of teacher's and student's prediction distributions. The closed-form solution of the optimization problem is:  
\begin{equation}
\label{teacher_Network}
t(Y) \propto s(Y|X)\text{exp}(CL(X,Y))~.
\end{equation}
The new objective which contains both ground truth labels and the teacher network is defined as:
\begin{equation}
\label{objective}
\min_{\phi \in \Phi } \frac{1}{n} \sum_{i=1}^{n}\alpha l(s_i,y_i)+(1-\alpha)l(s_i,t_i),
\end{equation}
where $s_i$ and $t_i$ are the student's and teacher's predictions for sample $i$; $y_i$ is the ground truth label for sample $i$; $\alpha$ is a balancing term between ground truth and the teacher network. $l$ is the loss function. More details can be found in \cite{eric1,eric2}.

\subsubsection{Knowledge Distillation for Visual Relationship Detection}
Linguistic knowledge is modeled by a conditional probability that encodes the strong correlation between the pair of objects \twotup{subj}{obj} and the predicate that humans tend to use to describe the relationship between them:
\begin{equation}
\label{language_knowledge}
L(X,Y)=\log P(pred|subj,obj),
\end{equation}
where X is the input data and $Y$ is the output distribution of the student network. $P(pred|subj,obj)$ is the conditional probability of a predicate given a fixed \twotup{subj}{obj} pair in the obtained linguistic knowledge set.

By solving the optimization problem in Eq.~\ref{optimization}, we construct a teacher network that is close to the student network, but penalizes a predicted predicate that is unlikely given the fixed \twotup{subj}{obj} pairs. The teacher's output can be viewed as a projection of the student's output in the solution space constrained by linguistic knowledge. For example, when predicting the predicate between a ``plate" and a ``table", given the subject (``plate") and the object (``table"), and the conditional probability $P(pred|plate,table)$, the teacher will penalize unlikely predicates, (\eg, ``in") and reward likely ones (\eg, ``on"), helping the network avoid portions of the parameter space that lead to poor solutions. 

Given the ground truth label and the teacher network's output distribution, we want the student network to not only predict the correct predicate labels but also mimic the linguistic knowledge regularized distributions. This is accomplished using a cross-entropy loss (see Eq.~\ref{objective}).

One advantage of this LK distillation framework is that it takes advantage of both knowledge-based and data-driven systems. Distillation works as a regularizer to help train the data-driven system. On the other hand, since we construct the teacher network based on the student network, the knowledge regularized predictions (teacher's output) will also be improved during training as the student's output improves. Rather than using linguistic knowledge as a post-processing step, our framework enables the data-driven model to absorb the linguistic knowledge together with the ground truth labels, allowing the deep network to learn a better visual model during training rather than only having its output modified in a post-processing step.
This leads to a data-driven model (the student) that generalizes better, especially in the zero-shot scenario where we lack linguistic knowledge about a \twotup{subj}{obj} pair. While \cite{hinton,eric1,eric2} used either the student or the teacher as the final output, our experiments show that both the student and teacher in our framework have their own advantages, so we combine them to achieve the best predictive power (see section \ref{exp}).



\subsubsection{Linguistic Knowledge Collection}
To obtain the linguistic knowledge $P(pred|subj,obj)$, a straightforward method is to count the statistics of the training annotations, which reflect the knowledge used by an annotator in choosing an appropriate predicate to describe a visual relationship. Due to the long tail distribution of relationships, a large number of combinations never occur in the training data; however, it is not reasonable to assume the probability of unseen relationships is 0. To tackle this problem, one can apply additive smoothing to assign a very small number to all 0's \cite{add_k}; however, the smoothed unseen conditional probabilities are uniform, which is still confusing at LK distillation time. To collect more useful linguistic knowledge of the long-tail unseen relationships, we exploit text data from the Internet.

One challenge of collecting linguistic knowledge online is that the probability of finding text data that specifically describes objects and their relationships is low. This requires us to obtain the knowledge from a huge corpus that covers a very large domain of knowledge. Thus we choose the Wikipedia 2014-06-16 dump containing around 4 billion words and 450 million sentences that have been parsed to text by \cite{wiki}\footnote{The Wikipedia text file can be found on \url{http://kopiwiki.dsd.sztaki.hu/}} to extract knowledge. 

We utilize the scene graph parser proposed in \cite{SGParser} to parse sentences into sets of \threetup{subj}{pred}{obj} triplets, and we compute the conditional probabilities of predicates based on these triplets.
However, due to the possible mistakes of the parser, especially on text from a much wider domain than the visual relationship detection task, the linguistic knowledge obtained can be very noisy. Naive methods such as using only the linguistic knowledge to predict the predicates or multiplying the conditional probability with the data-driven model's output fail. Fortunately, since the teacher network of our LK-distillation framework is constructed from the student network that is also supervised by the labeled data, a well-trained student network can help correct the errors from the noisy external probability. 
To achieve good predictive power on the seen and unseen relationships, we obtain the linguistic knowledge from both training data and the Wikipedia text corpus by a weighted average of their conditional probabilities when we construct the teachers' network, as shown in Eq.~\ref{objective}. We conduct a two-step knowledge distillation: during the first several training epoches, we only allow the student to absorb the knowledge from training annotations to first establish a good data-driven model. 
After that, we start distilling the external knowledge together with the knowledge extracted from training annotations weighted by the balancing term $C$ as shown in Eq.~\ref{objective}. 
The balancing terms are chosen by a validation set we select randomly from the training set (\eg, in VRD dataset, we select 1,000 out of 4,000 images to form the validation set) to achieve a balance between good generalization on the zero-shot and good predictive power on the entire testing set. 

\subsection{Semantic and Spatial Representations}
In \cite{VRD}, Lu \etal used the cropped image containing the union of two objects' bounding boxes to predict the predicate describing their relationship. While the cropped image encodes the visual appearance of both objects, it is difficult to directly model the strong semantic and spatial correlations between predicates and objects, as both semantic and spatial information is buried within the pixel values of the image. Meanwhile, the semantic and spatial representations capture similarities between visual relationships, which can generalize better to unseen relationships.

We utilize word-embedding \cite{word2vec} to represent the semantic meaning of each object by a vector. We then extract spatial features similarly to the ones in \cite{varun}:
\begin{equation}
\label{Spatial_Feature}
\left[\frac{x_{min}}{W},\frac{y_{min}}{H},\frac{x_{max}}{W},\frac{y_{max}}{H},\frac{A}{A_{img}}\right],
\end{equation}
where $W$ and $H$ are the width and height of the image, $A$ and $A_{img}$ are the areas of the object and the image, respectively. We concatenate the above features of two objects as the spatial feature (SF) for a \twotup{subj}{obj} pair. 

We predict the predicate conditioned on the semantic and spatial representations of the subject and object:
\begin{align}
\label{decompose}
P(R|I) =&P(pred|subj,obj,B_s,B_o,I)\nonumber \\
&~~ \cdot P(subj,B_s|I)P(obj,B_o|I),
\end{align}
where $subj$ and $obj$ are represented using the semantic object representation, $B_s$ and $B_o$ are the spatial features, and $I$ is the image region of the union of the two bounding boxes. 
For the BB-Union input, we use the same VGG-net \cite{VGG} in \cite{VRD} to learn the visual feature representation. 
We adopt a pre-trained word2vec vectors weighted by confidence scores of each object category for the subject and the object, then concatenate the two vectors as the semantic representation of the subject and the object.

%% file: Experiment.tex
\section{Experiments}\label{exp}
We evaluate our method on Visual Relationship Detection \cite{VRD} and Visual Genome \cite{VG} datasets for three tasks:
\textbf{Predicate detection}: given an input image and a set of ground truth bounding boxes with corresponding object categories, predict a set of predicates describing each pair of objects. This task evaluates the prediction of predicates without relying on object detection. 
\textbf{Phrase detection}: given an input image, output a phrase \threetup{subj}{pred}{obj} and localize the entire phrase as one bounding box.
\textbf{Relationship detection}: given an input image, output a relationship \threetup{subj}{pred}{obj} and both the subject and the object with their bounding boxes.


Both datasets have a zero-shot testing set that contains relationships that never occur in the training data. We evaluate on the zero-shot sets to demonstrate the generalization improvements brought by linguistic knowledge distillation.

\textbf{Implementation Details.} We use VGG-16 \cite{VGG} to learn the visual representations of the BB-Union of two objects. 
We use a pre-trained word2vec \cite{word2vec}  model to project the subjects and objects into vector space, and the final semantic representation is the weighted average based on the confidence scores of a detection. For the balancing terms, we choose $C=1$ and $\alpha=0.5$ to encourage the student network to mimic the teacher and the ground truth equally.

\textbf{{Evaluation Metric.}} We follow \cite{VRD,UIUC} using Recall@$n$ (R@$n$) as our evaluation metric (mAP metric would mistakenly penalize true positives because annotations are not exhaustive).
For two detected objects, multiple predicates are predicted with different confidences. The standard R@$n$ metric ranks all predictions for all object pairs in an image and compute the recall of top $n$. However, instead of computing recall based on all predictions, \cite{VRD} considers only the predicate with highest confidence for each object pair. 
Such evaluation is more efficient and forced the diversity of object pairs. 
However, multiple predicates can correctly describe the same object pair and the annotator only chooses one as ground truth, \eg, when describing a person ``next to" another person, predicate ``near" is also plausible.
So we believe that a good predicted distribution should have high probabilities for all plausible predicate(s) and probabilities close to 0 for remaining ones.  Evaluating only the top prediction per object pair may mistakenly penalize correct predictions since annotators have bias over several plausible predicates. 
So we treat the number of chosen predictions per object pair ($k$) as a hyper-parameter, and report R@$n$ for different $k$'s to compare with other methods \cite{VRD,UIUC,VisualPhrase}. Since the number of predicates is 70, $k=70$ is equivalent to evaluating all predictions \wrt two detected objects. 


\begin{table}[t]
\centering
\scriptsize
\setlength{\tabcolsep}{3.5pt} 
\caption{Predicate Detection on VRD Testing Set: ``U" is the union of two objects' bounding boxes; ``SF" is the spatial feature; ``W" is the word-embedding based semantic representations; ``L" means using LK distillation; ``S" is the student network; ``T" is the teacher network and ``S+T" is the combination of them. Part 1 uses the VRD training images; Part 2 uses the training images in VRD \cite{VRD} and images of Visual Genome (VG) \cite{VG} dataset.}
\label{P_VRD}
\begin{tabular}{@{}lccc|ccc@{}}
\toprule
\multicolumn{1}{c}{} & \multicolumn{3}{c}{Entire Set} & \multicolumn{3}{c}{Zero-shot} \\ \midrule
\multicolumn{1}{@{}l}{ } & R@100/50\tablefootnote{In predicate detection, R@100,k=1 and R@50,k=1 are equivalent (also observed in \cite{VRD}) because there are not enough objects in ground truth to produce over 50 pairs.}     & R@100    & R@50     & R@100/50     & R@100    & R@50    \\
\multicolumn{1}{c}{} & k=1      & k=70      & k=70     & k=1      & k=70      & k=70    \\\midrule
\multicolumn{3}{@{} l}{Part 1: Training images VRD only}&&&&\\
Visual Phrases \cite{VisualPhrase}       & 1.91             & -           & -    & - & - & -      \\
Joint CNN  \cite{Rcnn}           & 2.03             & -           & -    & - & - & -      \\

VRD-V only \cite{VRD}                                    & 7.11      & 37.20
\tablefootnote{\label{70}The recall of different k's are not reported in \cite{VRD}.We calculate those recall values using their code.}          
& 28.36       & 3.52      & 32.34           & 23.95   \\ 
VRD-Full \cite{VRD}                                & 47.87     & 84.34           & 70.97       & 8.45      & 50.04           & 29.77   \\\hdashline

Baseline: U only                                       & 34.82     & 83.15       & 70.02    & 12.75     & 69.42       & 47.84   \\
Baseline: L only                                       & 51.34     & 85.34       & 80.64    & 3.68      & 18.22       & 8.13   \\\hdashline
U+W                             & 37.15     & 83.78       & 70.75   & 13.44     & 69.77       & 49.01    \\
U+W+L:S               & 42.98     & 84.94       & 71.83  & 13.89       & 72.53       & 51.37     \\
U+W+L:T               & 52.96     & 88.98       & 83.26   & 7.81          & 40.15      & 32.62   \\\hdashline
U+SF                 &36.33           & 83.68       & 69.87  & 14.33     & 69.01       & 48.32    \\
U+SF+L:S                       & 41.06     & 84.81       & 71.27    & 15.14          & 72.72       & 51.62  \\
U+SF+L:T               & 51.67     & 87.71       & 83.84     & 8.05          & 41.51       & 32.77    \\ \hdashline
U+W+SF                   & 41.33     & 84.89       & 72.29  & 14.13         & 69.41       & 48.13    \\
U+W+SF+L: S            & 47.50      & 86.97       & 74.98  & \textbf{16.98}       & \textbf{74.65}       & \textbf{54.20}    \\
U+W+SF+L: T          & 54.13     & 89.41       & 82.54   & 8.80           & 41.53       & 32.81    \\
U+W+SF+L: T+S    & \textbf{55.16}     & \textbf{94.65}       & \textbf{85.64}    & -          & -       & -  \\
\midrule

\multicolumn{3}{@{} l}{Part 2: Training images VRD + VG}
                                              & & &  &                      \\
Baseline: U                                           & 36.97     & 84.49  & 70.19   & 13.31     & 70.56  & 50.34     \\
U+W+SF                     & 42.08     & 85.89       & 72.83   & 14.51          & 70.79       & 50.64     \\
U+W+SF+L: S             & 48.61      & 87.15       & 75.45    & 17.16          & 75.26       & 55.41     \\
U+W+SF+L: T             & 54.61     & 90.09       & 82.97       & 9.23          & 43.21       & 33.40  \\
U+W+SF+L: T+S               & 55.67     &95.19      & 86.14    & -          & -       & -    \\
\bottomrule
\end{tabular}
\end{table}

\begin{figure*}[t]
\centering     
\subfigure[Seen relationships]{\label{v}\includegraphics[width=140mm]{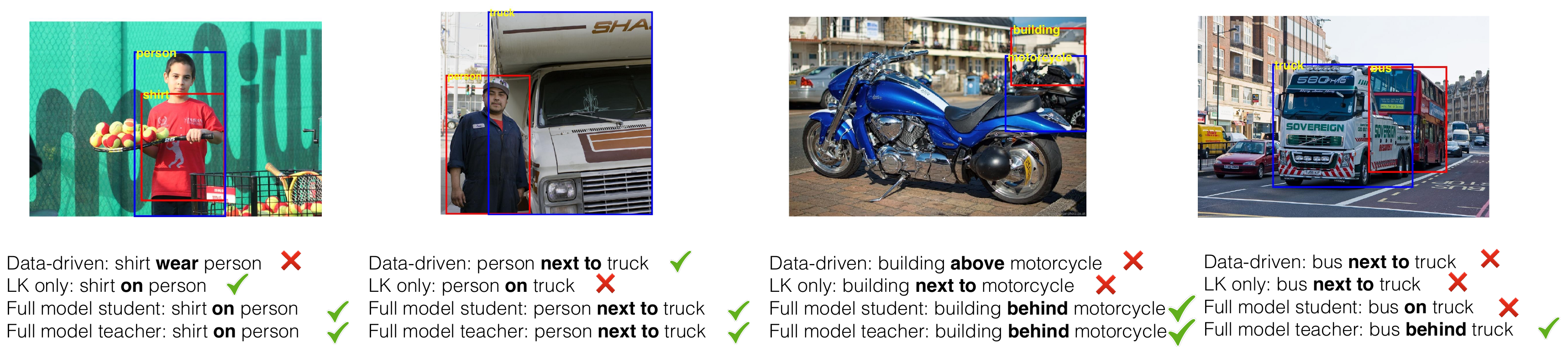}}

\subfigure[Zero-shot Relationships]{\label{vz}\includegraphics[width=140mm]{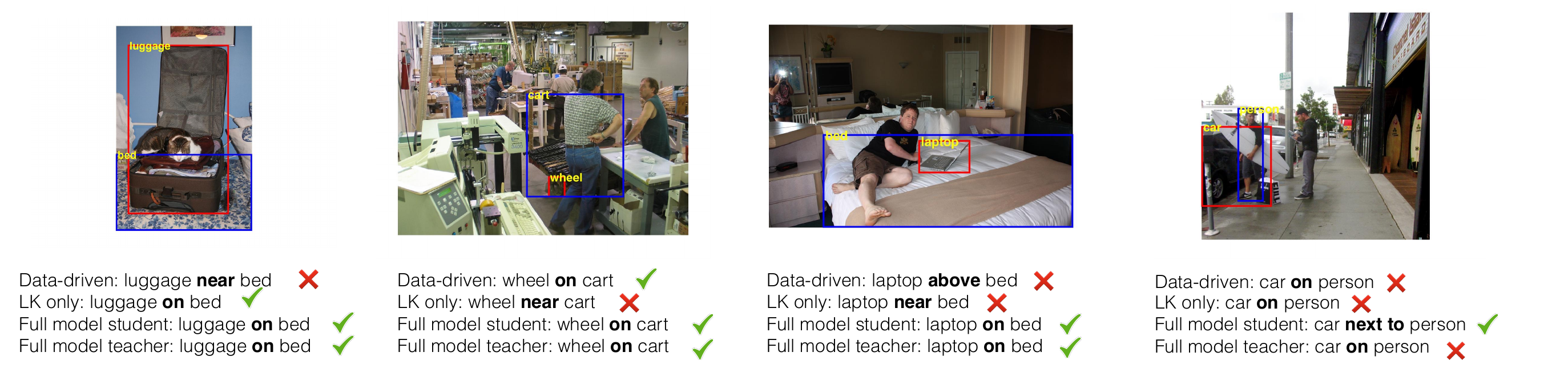}}
\caption{Visualization of predicate detection results: ``Data-driven" denotes the baseline using BB-Union; ``LK only" denotes the baseline using only the linguistic knowledge without looking at the image; ``Full model student" denotes the student network with U+W+SF features; ``Full model teacher" denotes the teacher network with U+W+SF features.}
\end{figure*}

\subsection{Evaluation on VRD Dataset}
\subsubsection{Predicate Prediction}
We first evaluate it on predicate prediction (as in \cite{VRD}).
Since \cite{UIUC,VIPCNN,RL} do not report results of predicate prediction, we compare our results with ones in \cite{VRD, VisualPhrase}. 

Part 1 of Table \ref{P_VRD} shows the results of linguistic knowledge distillation with different sets of features in our deep neural networks. In addition to the data-driven baseline ``Baseline: U only", we also compare with the baseline that only uses linguistic priors to predict a predicate, which is denoted as ``Baseline: L only". The ``Visual Phrases" method \cite{VisualPhrase} trains deformable parts models for each relationship; ``Joint CNN" \cite{Rcnn} trains a 270-way CNN to predict the subject, object and predicate together. The visual only model and the full model of \cite{VRD} that uses both visual input and language priors are denoted as ``VRD-V only" and ``VRD-Full". S denotes using the student network's output as the final prediction; T denotes using the teacher network's output. T+S denotes that for \twotup{subj}{obj} pairs that occur in the training data, we use the teacher network's output as the final prediction; for \twotup{subj}{obj} pairs that never occur in training, we use the student network's output. 

\textbf{End-to-end CNN training with semantic and spatial representations.} Comparing our baseline, which uses the same visual representation (BB-Union) as \cite{VRD}, and the ``VRD-V only" model, our huge recall improvement (R@100/50, k=1 increases from 7.11\% \cite{VRD} to 34.82\%) reveals that the end-to-end training with soft-max prediction outperforms extracting features from a fixed CNN + linear model method in \cite{VRD}, highlighting the importance of fine-tuning. In addition, adding the semantic representation and the spatial features improves the predictive power and generalization of the data-driven model\footnote{More analysis on using different combinations of features can be found in the supplementary materials.}.

To demonstrate the effectiveness of LK-distillation, we compare the results of using different combinations of features with/without using LK-distillation. In Part 1 of Table \ref{P_VRD}, we train and test our model on only the VRD dataset, and use the training annotation as our linguistic knowledge. 
\textit{``Linguistic knowledge only" baseline} (``Baseline: L only") itself has a strong predictive power and it outperforms the state-of-the-art method \cite{VRD} by a large margin (\eg, 51.34\%  vs. 47.87\% for R@100/50, k=1 on the entire VRD test set), which implies the knowledge we distill in the data-driven model is reliable and discriminative. 
However, since, some \twotup{subj}{obj} pairs in the zero-shot test set never occur in the linguistic knowledge extracted from the VRD train set, trusting only the linguistic knowledge without looking at the images leads to very poor performance on the zero-shot set of VRD, which explains the poor generalization of ``Baseline: L only" method and addresses the need for combining both data-driven and knowledge-based methods as the LK-distillation framework we propose does.

\begin{table*}[t]
\centering
\caption{Phrase and Relationship Detection: Distillation of Linguistic Knowledge. We use the same notations as in Table \ref{P_VRD}.}\vskip 3pt
\label{Detection}
\scriptsize
\setlength\tabcolsep{3.5pt}
\begin{tabular}{@{}lcccccc|cccccc@{}}
\toprule
& \multicolumn{6}{c|}{\bf Phrase Detection}                                               & \multicolumn{6}{|c}{\bf Relationship Detection}                                        \\ 
& R@100,                 & R@50,                  & R@100,                 & R@50,                  & R@100,                 & R@50,                  & R@100,                 & R@50,  & R@100,                 & R@50,                  & R@100,                 & R@50,                \\
& k=1                    & k=1                    & k=10                   & k=10                   & k=70                    & k=70                    & k=1                   & k=1     & k=10                   & k=10                    & k=70                   & k=70              \\\midrule
\multicolumn{3}{@{} l}{Part 1: Training images VRD only}&&&&\\
Visual Phrases \cite{VisualPhrase}       & 0.07             & 0.04           & -    & - & - & -   & -& -& -& -& -   \\
Joint CNN  \cite{Rcnn}           & 0.09             & 0.07           & -    & - & - & - & 0.09    & 0.07 & - & -& -    & -      \\
VRD - V only  \cite{VRD}           & 2.61             & 2.24           & -    & - & - & - & 1.85    & 1.58 & - & -& -    & -      \\
VRD - Full \cite{VRD}        & 17.03                  & 16.17                  & 25.52                     & 20.42       & 24.90                      & 20.04               & 14.70                  & 13.86                  & 22.03                      & 17.43            & 21.51                    & 17.35          \\
Linguistic Cues   \cite{UIUC}       & -                      & -                      & 20.70                  & 16.89                & --                      & --  & --                      & --                      & 18.37                  & 15.08         & --                      & --         \\
VIP-CNN   \cite{VIPCNN}       & \textbf{27.91}                      & 22.78                      & -                  & -                & --                      & --  & 20.01                      & 17.32                      & -                 & -         & --                      & --         \\
VRL   \cite{RL}       & 22.60                     & 21.37                      & -                 & -               & --                      & --  & 20.79                      & 18.19                      & -                  & -        & --                      & --         \\\hdashline
U+W+SF+L: S      & 19.98 & 19.15 & 25.16 & 22.95 & 25.54	& 22.59 & 17.69 & 16.57 & 27.98 & 19.92 &28.94	&20.12\\
U+W+SF+L: T      & 23.57 & 22.46 & 29.14 & 25.96 &29.09	&25.86 & 20.61 & 18.56 & 29.41 & 21.92 &31.13	&21.98\\
U+W+SF+L: T+S  & 24.03 & \textbf{23.14} & \textbf{29.76} & \textbf{26.47} &\bf29.43	&\bf26.32 & \textbf{21.34} & \textbf{19.17} & \textbf{29.89} & \textbf{22.56} &\bf 31.89	&\bf22.68\\

\midrule

\multicolumn{3}{@{} l}{Part 2: Training images VRD + VG}&&&&&&&&&&\\
U+W+SF+L: S      & 20.32 & 19.96 & 25.71 & 23.34 & 25.97	& 22.83 & 18.32 & 16.98 & 28.24 & 20.15 &29.85	&21.88\\
U+W+SF+L: T      & 23.89 & 22.92 & 29.82 & 26.34 &29.97	&26.15 & 20.94 & 18.93 & 29.95 & 22.62 &31.78	&22.65\\
U+W+SF+L: T+S     & 24.42 & 23.51 & 30.13 & 26.73 &30.01 &26.58 & 21.72 & 19.68 & 30.45 & 22.84 &32.56	&23.18\\ 
\bottomrule

\end{tabular}
\end{table*}

\begin{table*}[t]
\centering
\scriptsize
\caption{Phrase and Relationship Detection: Distillation of Linguistic Knowledge - Zero Shot. We use the same notations as in Table \ref{P_VRD}.}\vskip 3pt
\setlength\tabcolsep{3.8pt}
\label{Detection_Zero}
\begin{tabular}{@{}lcccccc|cccccc@{}}
\toprule
                   & \multicolumn{6}{c|}{\bf Phrase Detection}                                                            & \multicolumn{6}{|c}{\bf Relationship Detection}                                                      \\ 
                   & R@100,                & R@50,                 & R@100,                 & R@50,                  & R@100,                & R@50,                 & R@100,                 & R@50,    & R@100,                & R@50,                 & R@100,                 & R@50,              \\
                   & k=1                   & k=1                   & k=10                   & k=10                   & k=70                   & k=70                   & k=1                   & k=1   & k=10                   & k=10                   & k=70                   & k=70                \\\midrule
\multicolumn{3}{@{} l}{Part 1: Training images VRD only}&&&&&&&&&&\\

VRD - V only  \cite{VRD}           & 1.12             & 0.95           & -    & - & - & - & 0.78    & 0.67 & - & -& -    & -      \\
VRD - Full \cite{VRD}   & 3.75                  & 3.36                  & 12.57                      & 7.56    & 12.92                      & 7.96                  & 3.52                  & 3.13                  & 11.46                      & 7.01      & 11.70                      & 7.13                \\
Linguistic Cues  \cite{UIUC}   & -                     & -                     & 15.23                  & 10.86    & -                      & -              & -                     & -                     & 13.43                  & 9.67   & -                      & -                \\
VRL   \cite{RL}       & 10.31                    & 9.17                      & -                 & -               & --                      & --  & 8.52                      & 7.94                      & -                  & -        & --                      & --         \\
\hdashline
U+W+SF+L: S   & \textbf{10.89} & \textbf{10.44} & \textbf{17.24} & \textbf{13.01} &\bf 17.24	&\bf 12.96 & \textbf{9.14} & \textbf{8.89} & \textbf{16.15} & \textbf{12.31} &\bf 15.89	&\bf 12.02\\ 	
U+W+SF+L: T      & 6.71 & 6.54 & 11.27 & 9.45 &9.84	&7.86 & 6.44 & 6.07 & 9.71 & 7.82 &10.21	&8.75\\
\midrule
   
\multicolumn{3}{@{} l}{Part 2: Training images VRD + VG}&&&&&&&&&&\\
U+W+SF+L: S      & 11.23 & 10.87 & 17.89 & 13.53 & 17.88	& 13.41 & 9.75 & 9.41 & 16.81 & 12.72 &16.37	&12.29\\
U+W+SF+L: T      & 7.03 & 6.94 & 11.85 & 9.88 &10.12	&8.97 & 6.89 & 6.56 & 10.34 & 8.23 &10.53	&9.03\\

\bottomrule

\end{tabular}
\end{table*}

\textbf{The benefit of LK distillation} is visible across all feature settings: the data-driven neural networks that absorb linguistic knowledge (``student" with LK) outperform the data-driven models significantly (\eg, R@100/50, k=1 is improved from 37.15\% to 42.98\% for ``U+W" features on the entire VRD test set).
We also observe consistent improvement of the recall on the zero-shot test set of data-driven models that absorb the linguistic knowledge. The student networks with LK-distillation yield the best generalization, and outperform the data-driven baselines and knowledge only baselines by a large margin.

Unlike \cite{hinton,eric1,eric2}, where either the student or the teacher is the final output, we achieve better predictive power by combining both: we use the teacher network to predict the predicates whose \twotup{subj}{obj} pairs occur in the training data, and use the student network for the remaining. The setting ``U+W+SF+LK: T+S" performs the best. 
Fig. \ref{v} and \ref{vz} show a visualization of different methods.

\subsubsection{Phrase and Relationship Detection}
To enable fully automatic phrase and relationship detection, we train a Fast R-CNN detector \cite{fastRCNN} using VGG-16 for object detection. Given the confidence scores of detected each detected object, we use the weighed word2vec vectors as the semantic object representation, and extract spatial features from each detected bounding box pairs. We then use the pipeline in Fig. \ref{fig:system} to obtain the predicted predicate distribution for each pair of objects. According to Eq. \ref{decompose}, we use the product of the predicate distribution and the confidence scores of the subject and object as our final prediction results. We also adopt the triplet NMS in \cite{VIPCNN} to remove redundant detections. To compare with \cite{VRD}, we report R@n, k=1 for both phrase detection and relationship detection. For fair comparison with \cite{UIUC} (denoted as ``Linguistic Cues"), we choose k=10 as they did to report recall. In addition, we report the full recall measurement k=70. Evaluation results on the entire dataset and the zero-shot setting are shown in Part 1 of Tables \ref{Detection} and \ref{Detection_Zero}. Our method outperforms the state-of-the-art methods in \cite{VRD} and \cite{UIUC} significantly on both entire testing set and zero-shot setting. The observations about student and teacher networks are consistent with predicate prediction evaluation. 
We also compare our method with the very recently introduced ``VIP-CNN" in \cite{VIPCNN} and ``VRL" \cite{RL} and achieve better or comparable results. For phrase detection, we achieve better results than \cite{RL} and get similar result for R@50 to \cite{VIPCNN}. One possible reason that \cite{VIPCNN} gets better result for R@100 is that they jointly model the object and predicate detection while we use an off-the-shelf detector.
For relationship detection, we outperform both methods, especially on the zero-shot set.


\subsection{Evaluation on Visual Genome Dataset}
We also evaluate predicate detection on Visual Genome (VG) \cite{VG}, the largest dataset that has visual relationship annotations. We randomly split the VG dataset into training (88,077 images) and testing set (20,000 images) and select the relationships whose predicates and objects occur in the VRD dataset. We conduct a similar evaluation on the dataset (99,864 relationship instances in training and 19,754 in testing; 2,056 relationship test instances are never seen in training). We use the linguistic knowledge extracted from VG and report predicate prediction results in Table \ref{VG}. 

Not surprisingly, we observe similar behavior as on the VRD dataset---LK distillation regularizes the deep model and improves its generalization. We conduct another experiment in which images from Visual Genome dataset augment the training set of VRD and evaluate on the VRD test set. From the Part 2 of Tables \ref{P_VRD}, \ref{Detection} and \ref{Detection_Zero}, we observe that training with more data leads to only marginal improvement over almost all baselines and proposed methods. However, for all experimental settings, our LK distillation framework still brings significant improvements, and the combination of the teacher and student networks still yields the best performance. This reveals that incorporating additional knowledge is more beneficial than collecting more data\footnote{Details can be found in the supplementary materials.}. 

\begin{table}[t]
\centering
\scriptsize
\setlength{\tabcolsep}{3.6pt} 
\caption{Predicate Detection on Visual Genome Dataset. Notations are the same as in Table \ref{P_VRD}.}
\label{VG}
\begin{tabular}{@{}lccc|ccc@{}}
\toprule
\multicolumn{1}{c}{} & \multicolumn{3}{c}{Entire Set} & \multicolumn{3}{c}{Zero-shot} \\ \midrule
\multicolumn{1}{c}{} & R@100/50     & R@100    & R@50     & R@100/50     & R@100    & R@50    \\
\multicolumn{1}{c}{} & k=1      & k=70      & k=70     & k=1      & k=70      & k=70    \\\midrule
U                    & 37.81    & 82.05    & 81.41    & 7.54     & 81.00    & 65.22   \\
U+W+SF               & 40.92    & 86.81    & 84.92    & 8.66     & 82.50    & 67.72   \\
U+W+SF+L: S         & 49.88    & 91.25    & 88.14    & \textbf{11.28}    & \textbf{88.23}    & \textbf{72.96}   \\
U+W+SF+L: T         & 55.02    & 94.92    & 91.47    & 3.94     & 62.99    & 47.62   \\
U+W+SF+L: T+S       & \textbf{55.89}    & \textbf{95.68}    & \textbf{92.31}    & -        & -        & -       \\ \bottomrule
\end{tabular}
\end{table}

\begin{table}[t]
\centering
\scriptsize
\setlength{\tabcolsep}{3.6pt} 
\caption{Predicate Detection on VRD Testing Set: External Linguistic Knowledge. Part 1 uses the LK from VRD dataset; Part 2 uses the LK from VG dataset; Part 3 uses the LK from both VRD and VG dataset. Part 4 uses the LK from parsing Wikipedia text; Part 5 uses the LK from from both VRD dataset and Wikipedia. Notations are the same as as in Table \ref{P_VRD}. }
\label{VRD_use_VG_Stat}
\begin{tabular}{@{}lccc|ccc@{}}
\toprule
\multicolumn{1}{c}{} & \multicolumn{3}{c}{Entire Set} & \multicolumn{3}{c}{Zero-shot} \\ \midrule
\multicolumn{1}{l}{} & R@100/50     & R@100    & R@50     & R@100/50     & R@100    & R@50    \\
\multicolumn{1}{l}{} & k=1      & k=70      & k=70     & k=1      & k=70      & k=70    \\\midrule
Part 1 LK: VRD  & \multicolumn{3}{c|}{} \\
VRD-V only \cite{VRD}                                    & 7.11      & 37.20           & 28.36       & 3.52      & 32.34           & 23.95   \\ 
VRD-Full \cite{VRD}                                & 47.87     & 84.34           & 70.97       & 8.45      & 50.04           & 29.77   \\\hdashline
U+W+SF+L: S            & 47.50      & 86.97       & 74.98  & 16.98       & 74.65       & 54.20    \\
U+W+SF+L: T          & 54.13     & 89.41       & 82.54   & 8.80           & 41.53       & 32.81     \\\midrule
Part 2 LK: VG & \multicolumn{3}{c|}{} \\
U+W+SF+L: S         & 45.00   & 81.64    & 74.76    & 16.88   & 72.29    & 52.51   \\
U+W+SF+L: T         & 51.54    & 87.00   & 79.70    & 11.01    & 54.66    & 45.25   \\\midrule
\multicolumn{2}{@{} l}{Part 3 LK: VRD+VG} & \multicolumn{2}{c|}{} \\
U+W+SF+L: S         & 48.21    & 87.76   & 76.51    & 17.21    & 74.89   & 54.65   \\
U+W+SF+L: T         & \textbf{54.82}    & \textbf{90.63}    & \textbf{83.97}    & 12.32     & 47.22    & 38.24  \\\midrule

\multicolumn{2}{@{} l}{Part 4 LK: Wiki} & \multicolumn{2}{c|}{} \\
U+W+SF+L: S         & 36.05    & 77.88   & 68.16    & 11.80   & 64.24    & 49.19   \\
U+W+SF+L: T         & 30.41    & 69.86   & 60.25    & 11.12    & 63.58    & 44.65  \\\midrule

\multicolumn{2}{@{} l}{Part 5 LK: VRD+Wiki} & \multicolumn{2}{c|}{} \\
U+W+SF+L: S         & 48.94   & 87.11   & 77.79    & \textbf{19.17}    & \textbf{76.42}    & \textbf{56.81}   \\
U+W+SF+L: T         & 54.06    & 88.93   & 81.78    & 9.65     & 42.24   & 34.61  \\\bottomrule
\end{tabular}
\end{table}

\subsection{Distillation with External Knowledge}
The above experiments show the benefits of extracting linguistic knowledge from internal training annotations and distilling them in a data-driven model. 
However, training annotations only represent a small portion of all possible relationships and do not necessarily represent the real world distribution, which has a long tail.
For unseen long-tail relationships in the VRD dataset, we extract the linguistic knowledge from external sources: the Visual Genome annotations and Wikipedia, whose domain is much larger than any annotated dataset. 
In Table \ref{VRD_use_VG_Stat}, we show predicate detection results on the VRD test set using our linguistic knowledge distillation framework with different sources of knowledge. 
From Part 2 and Part 4 of Table \ref{VRD_use_VG_Stat}, we observe that using only the external knowledge, especially the very noisy one obtained from Wikipedia, leads to bad performance. 
However, interestingly, although the external knowledge can be very noisy (Wikipedia) and has a different distribution when compared with the VRD dataset (Visual Genome), the performance of the teacher network using external knowledge is much better than using only the internal knowledge (Part 1). 
This suggests that by properly distilling external knowledge, our framework obtains both good predictive power on the seen relationships and better generalization on unseen ones. 
Evaluation results of combining both internal and external linguistic knowledge are shown in Part 3 and Part 5 of Table \ref{VRD_use_VG_Stat}. We observe that by distilling external knowledge \textit{and} the internal one, we improve generalization significantly (\eg, LK from Wikipedia boosts the recall to 19.17\% on the zero-shot set) while maintaining good predictive power on the entire test set.

\begin{figure}[!t]
\centering
  \includegraphics[height=3cm]{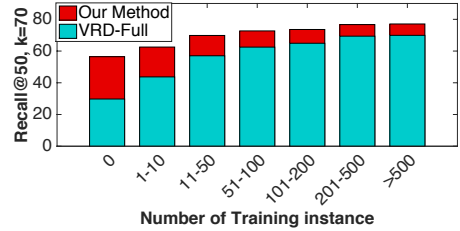}
  \caption{Performance with varying sizes of training examples. ``Our Method" denotes the student network that absorbs linguistic knowledge from both VRD training annotations and the Wikipedia text. ``VRD-Full" is the full model in \cite{VRD}.}
\label{fig:im}
\end{figure}


Fig. \ref{fig:im} shows the comparison between our student network that absorbs linguistic knowledge from both VRD training annotations and the Wikipedia text (denoted as ``Our Method") and the full model in \cite{VRD} (denoted as ``VRD-Full"). We observe that our method significantly outperforms the existing method, especially for the zero-shot (relationships with 0 training instance) and the few-shot setting (relationships with few training instances, \eg, $\le10$). By distilling linguistic knowledge into a deep model, our data-driven model improves dramatically, which is hard to achieve by only training on limited labeled images.
